# Parameters Optimization for Improving ASR Performance in Adverse Real World Noisy Environmental Conditions

**Urmila Shrawankar**                                                                                         *urmila@ieee.org*
IEEE Student Member,
*Computer Science and Engineering Department*
*G H Raisoni College of Engineering,*
*Nagpur, INDIA*

**Vilas Thakare**                                                                                            *vilthakare@yahoo.com*
Professor,
*PG Dept. of Computer Science,*
*SGB Amravati University,*
*Amravati, INDIA*

**Abstract**

From the existing research it has been observed that many techniques and methodologies are available for performing every step of Automatic Speech Recognition (ASR) system, but the performance (Minimization of Word Error Recognition-WER and Maximization of Word Accuracy Rate- WAR) of the methodology is not dependent on the only technique applied in that method. The research work indicates that, performance mainly depends on the category of the noise, the level of the noise and the variable size of the window, frame, frame overlap etc is considered in the existing methods.

The main aim of the work presented in this paper is to use variable size of parameters like window size, frame size and frame overlap percentage to observe the performance of algorithms for various categories of noise with different levels and also train the system for all size of parameters and category of real world noisy environment to improve the performance of the speech recognition system.

This paper presents the results of Signal-to-Noise Ratio (SNR) and Accuracy test by applying variable size of parameters. It is observed that, it is really very hard to evaluate test results and decide parameter size for ASR performance improvement for its resultant optimization.
Hence, this study further suggests the feasible and optimum parameter size using Fuzzy Inference System (FIS) for enhancing resultant accuracy in adverse real world noisy environmental conditions.

This work will be helpful to give discriminative training of ubiquitous ASR system for better Human Computer Interaction (HCI).

## 1. INTRODUCTION
Many Speech User Interface (SUI) based applications are now a part of daily life. However, a number of hurdles remain to making these technologies ubiquitous [1]. In light of the increasingly mobile and socially connected population, core challenges include robustness to additive background noise, convolutional channel noise, room reverberation and microphone mismatch [2, 3]. Other challenges include the ability to support the world's range of speakers, languages and dialects in speech technology.

Urmila Shrawankar & Vilas Thakare

Automated speech recognition (ASR) is the foundation of many speech and language processing applications. ASR technology includes signal processing, optimization, machine learning, and statistical techniques to model human speech and understanding.

This complete work focuses on following major issues for ASR performance improvement,
- Methodologies at pre-processing i.e. back-end level;
- Techniques at signal processing front-end for feature parameter extractions;
- Multi-environment training for Environment Adaptation and reducing the difference between training and testing environment;
- Variable parameter optimization using Fuzzy logic that is similar to the way of human thinking. Fuzzy sets are successfully applied for speech recognition due to their ability to deal with uncertainty.

This paper focuses on the last issue, as first three issues are already analyzed and results are submitted for publication.

This work may be extended to train the system for multi-user and English language speakers from various countries.

## 2. FUZZY LOGIC AND FUZZY INFERENCE METHODOLOGY

The concept of fuzzy logic [4] to present vagueness in linguistics, and further implement and express human knowledge and inference capability in a natural way. Fuzzy logic starts with the concept of a fuzzy set.

A fuzzy set is a set without a crisp, clearly defined boundary. It can contain elements with only a partial degree of membership. A Membership Function (MF) is a curve that defines how each point in the input space is mapped to a membership value (or degree of membership) between 0 and 1. The input space is sometimes referred to as the universe of discourse. Let X be the universe of discourse and x be a generic element of X. A classical set A is defined as a collection of elements or objects $x \in X$, such that each x can either belong to or not belong to the set A, $A \sqsubseteq X$. By defining a characteristic function (or membership function) on each element x in X, a classical set A can be represented by a set of ordered pairs (x, 0) or (x, 1), where 1 indicates membership and 0 non-membership. Unlike conventional set mentioned above fuzzy set expresses the degree to which an element belongs to a set. Hence the characteristic function of a fuzzy set is allowed to have value between 0 and 1, denoting the degree of membership of an element in a given set. If X is a collection of objects denoted generically by x, then a fuzzy set A in X is defined as a set of ordered pairs.

**The Fuzzy System has Five Parts of the Fuzzy Inference System**
- Fuzzification of the given set of variables
- Application of the fuzzy operator (AND or OR) in the antecedent
- Implication from the antecedent to the consequent
- Aggregation of the consequents across the rules
- Defuzzification

**Fuzzy Inference System**
In this context, Fuzzy Inference Systems (FIS), also known as fuzzy rule-based systems, are well-known tools for the simulation of nonlinear behaviors with the help of fuzzy logic and linguistic fuzzy rules. There are some popular inference techniques developed for fuzzy systems, such as Mamdani [5], Sugeno [6], Tsukamoto [6]. Mamdani FIS is selected to use in this experimental study.

## 3. PROPOSED METHODOLOGY

From the literature study and analysis of speech processing methods it is observed that performance of the speech processing technique and the word recognition accuracy of a speech recognition system is dependent on windowing and frame size frame overlap size of a speech sample [7], recoding – training – testing environment, technique/s used at front-end and back-end of a system.

Urmila Shrawankar & Vilas Thakare

Therefore this work uses variable size of windowing, framing and frame overlap size, and the performance evaluation is done on every step of a system model from front-end and back-end techniques.

- Speech samples of digits, zero to nine are recorded from different ten Indian English speaking persons (five males and five females) and multiple utterances, in real world noisy environment with sampling frequency 8 kHz and time duration 3 sec.
- First, these samples are checked for whether voiced / invoiced / or silence [8]. Only voiced samples are considered and others are discarded.
- In the pre-processing steps, noise is removed using filters and enhanced [9,10] using Wiener-Type Filter algorithm [11]. This algorithm is tested on different window size, frame size frame overlap size and for different category of noisy environment (Back-end level).
- SNR improvement test is performed. Results are given in Table: 1-5.
- Features are extracted using MFCC front-end technique [12, 13]. Features are extracted using different window and frame size.
- Further these feature parameters are passed to Hidden Markov Model (HMM) for training and followed by recognition [14]. Here the aim is to train the system for all types of environment (Multi-environment training) to improve the word recognition accuracy therefore, system is trained for all variety of samples like samples recorded at clean environment (inside glass cabin), samples recorded at all category of real world noise (out-side of room and at crowded places), samples after applying traditional noise removal filters, samples after applying speech enhancement algorithms etc.
- Accuracy is computed using Word recognition rate separately for different window and frame size. Results are given in Table: 1-5.
- This experiment is performed adjusting variable parameters like window, frame and frame overlap size manually (using computer program) to find out improvement in word recognition accuracy using iterative method. Please refer Table: 1-5
- The aim of this experiment is to find-out variable parameters size to optimized accuracy therefore a ruled base Fuzzy Inference System (FIS) from MatLab [15] is used.
- Window size and Frame overlap size in % and SNR as an environment are sent to the FIS as input parameters and Word recognition accuracy is computed as output. Rules are framed to compute the output.

## 4. EMPIRICAL PROCESS FOR FUZZY INFERENCE SYSTEM (FIS)
FIS uses following parameters,

**4.1 Parameter List:**

1. Hamming Window Size: 240-270 step size 10 (240, 250, 260, 270)

2. Frame Overlap percentage: 20-60 % Step size 5% (20, 25, 30, 35, 40, 45, 50, 55, 60)

3. Window Size is calculated using following equation:

   Window Size = Window length * Sampling Frequency (Window length is 20 ms)

4. Variable Frame Size is obtained using equation:

$$\text{Frame Size} = \frac{\text{Speech Sample Length}}{\text{Size of Hamming Window}} * \text{Frame Overlap \%}$$

Urmila Shrawankar & Vilas Thakare

5. Word Recognition Accuracy is computed using equation:

$$\text{Word Recognition Accuracy} = \frac{\text{Number of Words Recognised}}{\text{Number of Words Tested}} \%$$

**4.2 Fuzzy Inference System (FIS) :**
FIS is set using following parameters:

**[System]**
Name='SpeechAccuracy'
Type='mamdani'
Version=2.0
NumInputs=3
NumOutputs=1
NumRules=5
AndMethod='min'
OrMethod='max'
ImpMethod='min'
AggMethod='max'
DefuzzMethod='centroid'

Three inputs are selected in the system, SNR value is passed for the Environment, Hamming windows size as WinSz and Frame overlap percentages as FrOver.
Input parameters, their membership function and ranges as follow.

**[Input1]**
Name='Environment'
Range=[10 50]
NumMFs=3
MF1='VNoisy':'trimf',[-6 10 20]
MF2='Noisy':'trimf',[20 30 35]
MF3='Clean':'trimf',[35 50 66]
Environment is defined as the value based on SNR, 10-20 dB is Very Noisy, 20-35 dB is Noisy and 35-50 dB is assumed for clean environment.

**[Input2]**
Name='WinSz'
Range=[240 270]
NumMFs=3
MF1='Small':'trimf',[225 240 250]
MF2='Medium':'trimf',[250 255 260]
MF3='Large':'trimf',[260 270 282]
Window size is considered in three ranges Small, Medium and Large with ranges 240-250, 255-260 and 260-270 respectively.

**[Input3]**
Name='FrOver'
Range=[20 60]
NumMFs=3
MF1='Small':'trimf',[4 20 40]
MF2='Medium':'trimf',[40 50 55]
MF3='Large':'trimf',[50 60 76]
Frame overlap percentage is considered in three ranges Small, Medium and Large with ranges 20-40, 40-50 and 50-60 respectively.

Urmila Shrawankar & Vilas Thakare

**[Output1]**
Name='Accuracy'
Range=[95 100]
NumMFs=3
MF1='Good':'gaussmf',[0.8493 95]
MF2='Better':'gaussmf',[0.8493 97.5]
MF3='Best':'gaussmf',[0.8493 100]

The Word recognition Accuracy is the final output. It is considered as Good, Better and Best in the expected range of 95 to 100 %,

After defining input, output and their membership functions, rules are framed and weights are assigned as given below

**[Rules]**
3 0 0, 2 (0.5) : 1
3 0 2, 3 (0.75) : 1
3 2 2, 3 (1) : 1
0 0 2, 2 (0.5) : 1
0 2 0, 2 (0.5) : 1

- If (Environment is Clean) then (Accuracy is Better) (0.5)
- If (Environment is Clean) and (FrOver is Medium) then (Accuracy is Best) (0.75)
- If (Environment is Clean) and (WinSz is Medium) and (FrOver is Medium) then (Accuracy is Best) (1)
- If (FrOver is Medium) then (Accuracy is Better) (0.5)
- If (WinSz is Medium) then (Accuracy is Better) (0.5)

Final step is defuzzification, output accuracy is observed for different rules and crisp value is obtained using centroid - DefuzzMethod,

Observations and output results are given in Results and Discussion section.

## 5. RESULTS AND DISCUSSION
Frame size, SNR and accuracy results for different Hamming window and frame overlap % are given in table 1-5. Tables are given at the end of paper.

**Table 1:** SNR & Accuracy Test results for different Hamming Window Size, Frame Size and Frame Overlap % for same sample recorded at Real World Environment Noise

**Table 2:** SNR & Accuracy Test results for different frame size and frame overlap % and Window Size 240 for different samples at Real World Environment Noise

**Table 3:** SNR & Accuracy Test results for different frame size and frame overlap % and Window Size 250 for different samples at Real World Environment Noise

**Table 4:** SNR & Accuracy Test results for different frame size and frame overlap % and Window Size 260 for different samples at Real World Environment Noise

**Table 5:** SNR & Accuracy Test results for different frame size and frame overlap % and Window Size 270 for different samples at Real World Environment Noise

Urmila Shrawankar & Vilas Thakare

**FIS Results**
- Five rules are set to compute the Accuracy as an output as shown in fig: 1.
- Using the default values output of rules are viewed as shown in fig: 2 and crisp value of accuracy is observed.

Output of rules are viewed and crisp value of accuracy is observed by changing input values as shown in fig: 3

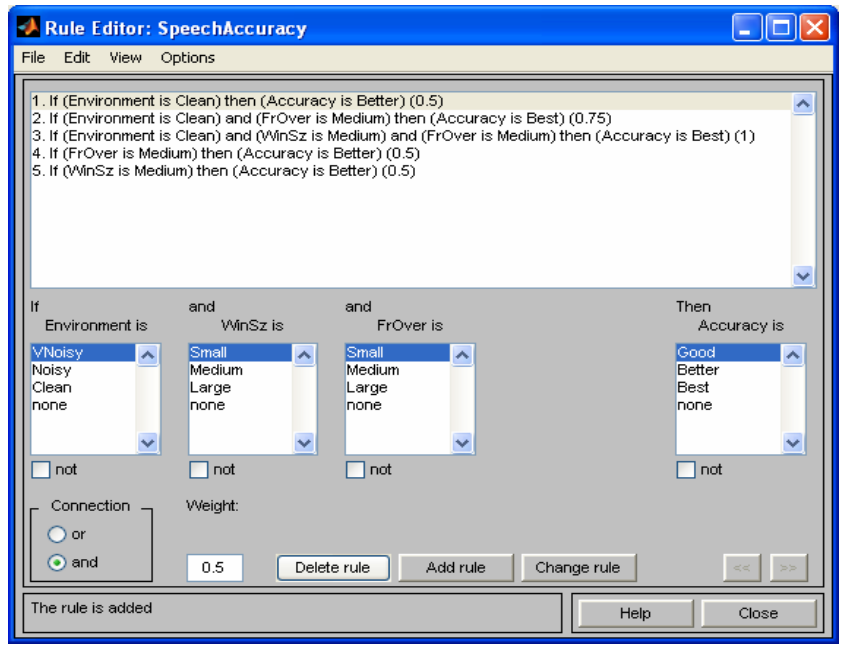

**Fig 1:** Rules sets for Accuracy Optimization

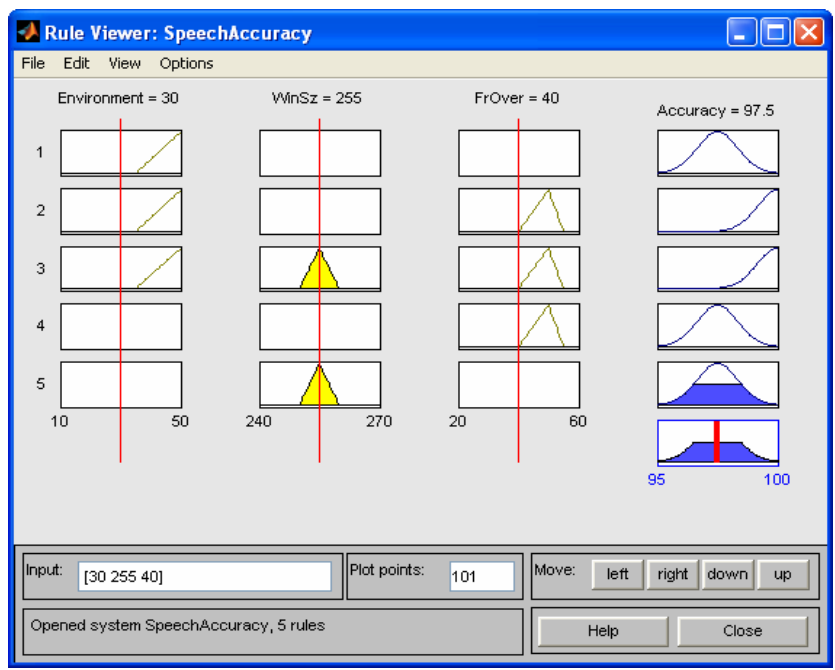

**Fig 2:** Output of Rules and Defuzzification (Parameter Set 1)

Urmila Shrawankar & Vilas Thakare

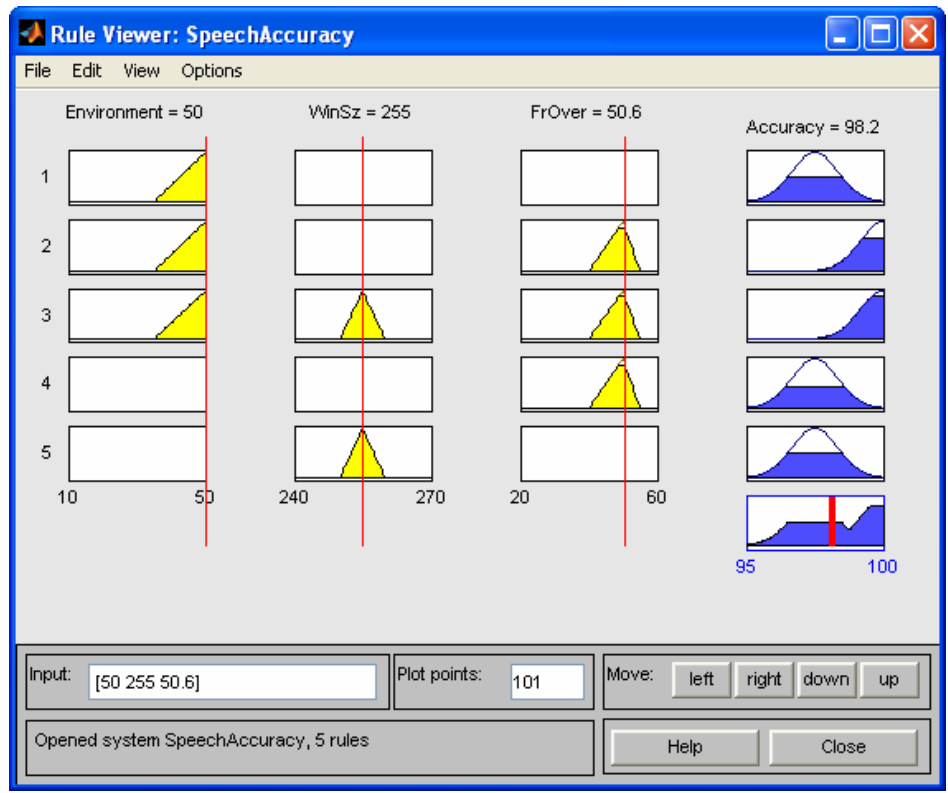

**Fig 3:** Output of Rules and Defuzzification (Parameter Set 2)

## 6. CONCLUSION
The assumption for this study was that the word recognition accuracy not only depends on the adverse environment conditions but variable size of hamming window, frame overlap and frame length also. It is proved by using traditional algorithm methods and calculations using different size of parameters as well as fuzzy system.

The improved word recognition accuracy is observed using hybrid signal enhancement method as compared to results shown in previous literature.

From the tabular data, for all hamming window size, SNR gradually improved till 50 % frame overlap but after going down. There is variation in word recognition accuracy calculated for different hamming window size and frame size. The better accuracy is observed in between 45-55 % frame overlap.

From FIS simulation results, the feasible parameter size for accuracy improvement is found in ranges, that clean environment SNR between 40-50 dB, Hamming window size should be medium 250-260 ms and frame overlap percentage between 40-55 %.

The optimized parameter size for best accuracy is observed by clean environment SNR above 45 db, hamming window size 255 ms and frame overlap percentage 50.6

**Result Tables**

| Hamm Win Size | Variables | Frame Overlap 20 % | Frame Overlap 25 % | Frame Overlap 30 % | Frame Overlap 35 % | Frame Overlap 40 % | Frame Overlap 45 % | Frame Overlap 50 % | Frame Overlap 55 % | Frame Overlap 60 % |
|---|---|---|---|---|---|---|---|---|---|---|
| 240 | FrSz | 11.0000 | 11.0000 | 12.0000 | 13.0000 | 13.0000 | 13.0000 | 14.0000 | 15.0000 | 15.0000 |
|  | SNR | 10.6244 | 13.9512 | 18.1852 | 21.9053 | 29.0342 | 37.6815 | 42.9845 | 33.3873 | 25.5868 |
|  | Accuracy | 95.6196 | 97.6584 | 96.3816 | 92.0023 | 95.8129 | 96.0878 | 97.1450 | 96.8232 | 97.2298 |
| 245 | FrSz | 11.0000 | 11.0000 | 12.0000 | 12.0000 | 13.0000 | 13.0000 | 14.0000 | 14.0000 | 15.0000 |
|  | SNR | 09.6427 | 13.1146 | 17.7529 | 22.2198 | 29.5783 | 39.1633 | 43.8332 | 33.0721 | 23.8667 |
|  | Accuracy | 86.7843 | 91.8022 | 94.0904 | 93.3232 | 97.6084 | 99.8664 | 99.0630 | 95.9091 | 90.6935 |
| 250 | FrSz | 11.0000 | 11.0000 | 12.0000 | 12.0000 | 13.0000 | 13.0000 | 14.0000 | 14.0000 | 15.0000 |
|  | SNR | 10.1832 | 13.9075 | 17.6812 | 23.4952 | 29.5787 | 38.8426 | 43.4540 | 33.3081 | 23.5664 |
|  | Accuracy | 91.6488 | 97.3525 | 93.7104 | 98.6798 | 97.6097 | 99.0486 | 98.2060 | 96.5935 | 89.5523 |
| 255 | FrSz | 10.0000 | 11.0000 | 11.0000 | 12.0000 | 12.0000 | 13.0000 | 13.0000 | 14.0000 | 14.0000 |
|  | SNR | 10.0862 | 13.9694 | 16.9383 | 22.2111 | 29.0016 | 36.1081 | 39.3579 | 30.8962 | 23.6862 |
|  | Accuracy | 90.7758 | 97.7858 | 89.7730 | 93.2866 | 95.7053 | 92.0757 | 98.9489 | 98.5990 | 90.0076 |
| 260 | FrSz | 10.0000 | 11.0000 | 11.0000 | 12.0000 | 12.0000 | 13.0000 | 13.0000 | 13.0000 | 14.0000 |
|  | SNR | 10.7480 | 13.5735 | 18.0468 | 22.0750 | 29.0213 | 38.0217 | 43.3384 | 33.5192 | 25.9135 |
|  | Accuracy | 96.7320 | 95.0145 | 95.6480 | 92.7150 | 95.7703 | 96.9553 | 97.9448 | 97.2057 | 98.4713 |
| 265 | FrSz | 10.0000 | 11.0000 | 11.0000 | 11.0000 | 12.0000 | 12.0000 | 13.0000 | 13.0000 | 14.0000 |
|  | SNR | 10.3382 | 13.7806 | 16.8835 | 21.9894 | 28.9723 | 37.2079 | 41.9009 | 33.2083 | 25.5744 |
|  | Accuracy | 93.0438 | 96.4642 | 89.4826 | 92.3555 | 95.6086 | 97.8801 | 98.6960 | 96.3041 | 97.1827 |
| 270 | FrSz | 10.0000 | 10.0000 | 11.0000 | 11.0000 | 12.0000 | 12.0000 | 13.0000 | 13.0000 | 14.0000 |
|  | SNR | 10.2738 | 13.6145 | 16.8504 | 22.0041 | 28.3733 | 36.2857 | 40.0923 | 32.2709 | 24.3966 |
|  | Accuracy | 92.4642 | 95.3015 | 89.3071 | 92.4172 | 93.6319 | 98.5285 | 98.6086 | 98.5856 | 92.7071 |

**TABLE 1:** SNR & Accuracy Test results for different Hamming Window Size, Frame Size and Frame Overlap % for same sample recorded at Real World Environment Noise

| Digit & SNR | Frame Overlap % | 20 | 25 | 30 | 35 | 40 | 45 | 50 | 55 | 60 |
|---|---|---|---|---|---|---|---|---|---|---|
| Zero 2.1552 | Fr Size | 13.1950 | 13.7708 | 14.4029 | 14.8725 | 15.5108 | 16.1554 | 16.6812 | 17.1727 | 17.7267 |
|  | SNR | 10.3807 | 14.1610 | 17.6800 | 23.0630 | 30.9840 | 38.9664 | 39.5634 | 29.5456 | 23.0126 |
|  | Accuracy | 83.0456 | 86.3821 | 86.6320 | 92.2520 | 98.2193 | 98.6227 | 99.1268 | 97.7731 | 82.8454 |
| One 2.532 | Fr Size | 11.4950 | 11.9479 | 12.4529 | 13.1006 | 13.6442 | 13.9804 | 14.4937 | 15.0415 | 15.3267 |
|  | SNR | 10.6244 | 13.9512 | 18.1852 | 21.9053 | 29.0342 | 37.6815 | 42.9845 | 33.3873 | 25.5868 |
|  | Accuracy | 84.9952 | 85.1023 | 89.1075 | 87.6212 | 92.0384 | 98.6675 | 98.9690 | 97.1457 | 92.1125 |
| Two 7.1607 | Fr Size | 12.2950 | 12.8594 | 13.4279 | 14.0850 | 14.3442 | 15.0679 | 15.7438 | 18.2383 | 18.9267 |
|  | SNR | 10.9099 | 14.9720 | 18.8613 | 22.4832 | 28.8421 | 37.4938 | 39.5090 | 31.8631 | 24.0928 |
|  | Accuracy | 87.2792 | 91.3292 | 92.4204 | 89.9328 | 91.4295 | 98.2357 | 99.0180 | 98.0304 | 86.7341 |
| Three 5.1581 | Fr Size | 11.6950 | 12.0781 | 12.6154 | 13.2975 | 13.6442 | 14.2523 | 16.6812 | 17.1727 | 18.1267 |
|  | SNR | 11.0012 | 14.1745 | 17.6697 | 22.8366 | 29.3103 | 39.6793 | 37.2644 | 30.2142 | 21.6886 |
|  | Accuracy | 88.0096 | 86.4645 | 86.5815 | 91.3464 | 92.9137 | 97.2624 | 99.5288 | 98.5783 | 78.0790 |
| Four 3.3196 | Fr Size | 8.8950 | 9.3438 | 9.6904 | 10.1475 | 13.1775 | 12.3492 | 13.2438 | 13.9758 | 14.5267 |
|  | SNR | 10.8110 | 13.6829 | 17.4169 | 22.2642 | 29.6222 | 41.9072 | 48.0202 | 33.8526 | 25.2955 |
|  | Accuracy | 86.4880 | 83.4657 | 85.3428 | 89.0568 | 93.9024 | 96.3866 | 99.0404 | 98.4020 | 91.0638 |
| Five 2.9423 | Fr Size | 11.7950 | 12.4688 | 12.7779 | 13.4944 | 13.8775 | 14.5242 | 16.9938 | 17.5279 | 18.1267 |
|  | SNR | 11.2457 | 14.5056 | 20.0666 | 22.6091 | 28.5618 | 37.8589 | 42.8678 | 33.9163 | 25.3535 |
|  | Accuracy | 89.9656 | 88.4842 | 98.3263 | 90.4364 | 90.5409 | 97.0755 | 98.7356 | 96.5740 | 91.2726 |
| Six 2.9731 | Fr Size | 7.5950 | 7.9115 | 8.3904 | 9.9506 | 10.3775 | 10.7179 | 11.0563 | 11.1342 | 12.1267 |
|  | SNR | 9.8293 | 13.1812 | 17.1789 | 22.1467 | 29.4873 | 41.3465 | 48.7289 | 32.2793 | 24.2792 |
|  | Accuracy | 78.6344 | 80.4053 | 84.1766 | 88.5868 | 93.4747 | 98.0970 | 99.4578 | 98.1541 | 87.4051 |
| Seven 3.963 | Fr Size | 14.8950 | 15.5938 | 16.1904 | 16.8412 | 17.6108 | 18.0585 | 18.8688 | 19.6592 | 20.1267 |
|  | SNR | 10.5861 | 14.1139 | 17.9920 | 22.6423 | 29.5017 | 41.1010 | 45.9814 | 35.6063 | 27.6307 |
|  | Accuracy | 84.6888 | 86.0948 | 88.1608 | 90.5692 | 93.5204 | 97.5323 | 98.9628 | 96.1370 | 91.4705 |
| Eight 4.0143 | Fr Size | 8.6723 | 8.9856 | 9.3950 | 9.7304 | 12.3792 | 12.1521 | 12.8019 | 13.2287 | 14.1477 |
|  | SNR | 10.5210 | 13.5585 | 17.4763 | 21.1063 | 26.8338 | 31.0465 | 30.3656 | 23.1879 | 22.1511 |
|  | Accuracy | 94.1680 | 92.7069 | 95.6339 | 94.4252 | 95.0631 | 97.4070 | 98.7312 | 96.6073 | 97.7440 |
| Nine 5.2752 | Fr Size | 13.8950 | 14.4219 | 15.2154 | 15.8569 | 16.4442 | 16.9710 | 17.3062 | 18.2383 | 21.3267 |
|  | SNR | 9.9747 | 13.9209 | 17.3987 | 22.2017 | 28.8799 | 36.1080 | 35.2710 | 28.7752 | 18.3880 |
|  | Accuracy | 97.7976 | 94.9175 | 95.2536 | 98.8068 | 93.5493 | 98.0484 | 99.5420 | 97.6930 | 96.1968 |

**TABLE 2:** SNR & Accuracy Test results for different frame size and frame overlap % and Window Size 240 for different samples at Real World Environment Noise

| Digit & SNR | Frame Overlap % | 20 | 25 | 30 | 35 | 40 | 45 | 50 | 55 | 60 |
|---|---|---|---|---|---|---|---|---|---|---|
| Zero 2.1552 | Fr Size | 12.7632 | 13.3450 | 13.8268 | 14.4666 | 14.8904 | 15.7702 | 16.0140 | 16.4858 | 17.0176 |
|  | SNR | 10.4638 | 14.8782 | 18.4936 | 23.4261 | 30.2385 | 40.6107 | 42.3616 | 28.6485 | 23.1388 |
|  | Accuracy | 83.7104 | 92.2448 | 90.6186 | 93.7044 | 95.8560 | 96.4046 | 98.3483 | 97.3510 | 83.2997 |
| One 2.532 | Fr Size | 11.1312 | 11.5950 | 12.1108 | 12.5766 | 13.0984 | 13.6822 | 14.2140 | 14.4398 | 15.0976 |
|  | SNR | 10.1832 | 13.9075 | 17.6812 | 23.4952 | 29.5787 | 38.8426 | 43.4540 | 33.3081 | 23.5664 |
|  | Accuracy | 81.4656 | 86.2265 | 86.6379 | 93.9808 | 93.7645 | 98.3380 | 99.7297 | 98.9319 | 84.8390 |
| Two 7.1607 | Fr Size | 11.8992 | 12.4700 | 12.8908 | 13.5216 | 13.9944 | 14.4652 | 15.1140 | 17.5088 | 18.1696 |
|  | SNR | 11.4743 | 14.8349 | 18.9242 | 22.7534 | 29.0030 | 40.5465 | 41.8622 | 33.1496 | 24.3065 |
|  | Accuracy | 91.7944 | 91.9764 | 92.7286 | 91.0136 | 91.9395 | 97.2570 | 98.2596 | 98.5039 | 87.5034 |
| Three 5.1581 | Fr Size | 11.2272 | 11.8450 | 12.2668 | 12.7656 | 13.0984 | 13.6822 | 16.0140 | 16.8268 | 17.4016 |
|  | SNR | 11.5551 | 13.4172 | 18.0454 | 22.9101 | 29.3780 | 40.4492 | 35.3010 | 28.0499 | 21.1498 |
|  | Accuracy | 92.4408 | 83.1866 | 88.4225 | 91.6404 | 93.1283 | 97.0332 | 98.9562 | 97.7347 | 76.1393 |
| Four 3.3196 | Fr Size | 8.8272 | 9.2200 | 9.6148 | 9.9306 | 12.6504 | 12.1162 | 13.0140 | 13.7578 | 14.3296 |
|  | SNR | 9.8627 | 13.5182 | 17.7196 | 22.7193 | 29.6216 | 42.1054 | 44.2421 | 33.4859 | 24.3936 |
|  | Accuracy | 78.9016 | 83.8128 | 86.8260 | 90.8772 | 93.9005 | 96.8424 | 98.4478 | 97.4119 | 87.8170 |
| Five 2.9423 | Fr Size | 11.5152 | 11.9700 | 12.5788 | 12.9546 | 13.5464 | 13.9432 | 16.3140 | 17.1678 | 18.1696 |
|  | SNR | 11.3458 | 14.4631 | 18.7603 | 23.4329 | 29.5343 | 39.1727 | 35.4104 | 32.8159 | 25.1857 |
|  | Accuracy | 90.7664 | 89.6712 | 91.9255 | 93.7316 | 93.6237 | 97.0972 | 98.1947 | 97.6029 | 90.6685 |
| Six 2.9731 | Fr Size | 7.4832 | 7.8450 | 8.0548 | 9.5526 | 10.1864 | 10.5502 | 10.9140 | 11.0298 | 11.6416 |
|  | SNR | 10.1006 | 14.4536 | 18.7031 | 22.5996 | 29.3225 | 39.6375 | 40.2627 | 31.3231 | 22.4486 |
|  | Accuracy | 80.8048 | 89.6123 | 91.6452 | 90.3984 | 92.9523 | 97.1663 | 98.7727 | 98.5724 | 80.8150 |
| Seven 3.963 | Fr Size | 14.4912 | 15.0950 | 15.6988 | 16.3566 | 16.9064 | 17.5972 | 18.1140 | 18.8728 | 19.7056 |
|  | SNR | 9.7025 | 12.8357 | 17.2056 | 22.5191 | 29.3951 | 41.3525 | 45.8197 | 36.3344 | 26.8617 |
|  | Accuracy | 77.6200 | 79.5813 | 84.3074 | 90.0764 | 93.1825 | 95.1108 | 99.8869 | 98.1029 | 96.7021 |
| Eight 7.5287 | Fr Size | 4.8912 | 5.8450 | 6.1828 | 6.5286 | 7.2744 | 6.1132 | 6.4140 | 12.3938 | 8.9536 |
|  | SNR | 10.3578 | 13.3700 | 17.4085 | 20.1015 | 23.0182 | 20.6113 | 17.3084 | 26.2883 | 20.3808 |
|  | Accuracy | 92.8624 | 92.8940 | 95.3017 | 90.4060 | 92.9677 | 97.4060 | 98.7323 | 97.9784 | 93.3709 |
| Nine 5.2752 | Fr Size | 13.4352 | 14.0950 | 14.6068 | 15.2226 | 15.7864 | 16.2922 | 17.2140 | 17.5088 | 20.4736 |
|  | SNR | 10.0927 | 13.7013 | 17.3809 | 22.7023 | 28.8386 | 38.8475 | 33.3886 | 26.0706 | 20.1303 |
|  | Accuracy | 90.7416 | 94.9481 | 85.1664 | 90.8092 | 91.4184 | 98.3493 | 98.7871 | 98.3906 | 92.4691 |

**TABLE 3:** SNR & Accuracy Test results for different frame size and frame overlap % and Window Size 250 for different samples at Real World Environment Noise

Urmila Shrawankar & Vilas Thakare

| Digit & SNR | Frame Overlap % | 20 | 25 | 30 | 35 | 40 | 45 | 50 | 55 | 60 |
|---|---|---|---|---|---|---|---|---|---|---|
| Zero 2.1552 | Fr Size | 12.2723 | 12.8317 | 13.2950 | 13.9102 | 14.3177 | 15.1637 | 15.3981 | 16.1796 | 16.3631 |
| | SNR | 13.0516 | 16.1677 | 17.3721 | 22.8980 | 30.9635 | 40.0116 | 35.2423 | 28.8058 | 22.6986 |
| | Accuracy | 91.3612 | 98.6230 | 85.1233 | 91.5920 | 98.1543 | 98.4274 | 98.9604 | 98.6562 | 81.7150 |
| One 2.532 | Fr Size | 10.7031 | 11.2692 | 11.6450 | 12.0929 | 12.5946 | 13.1560 | 13.6673 | 13.8844 | 14.5169 |
| | SNR | 10.7480 | 13.5735 | 18.0468 | 22.0750 | 29.0213 | 38.0217 | 43.3384 | 33.5192 | 25.9135 |
| | Accuracy | 75.2360 | 82.7984 | 88.4293 | 88.3000 | 91.9975 | 98.7312 | 99.3430 | 97.8538 | 93.2886 |
| Two 7.1607 | Fr Size | 11.4415 | 11.9904 | 12.3950 | 13.0015 | 13.4562 | 13.9088 | 14.5327 | 16.8354 | 17.4708 |
| | SNR | 10.1636 | 13.7098 | 17.8231 | 22.2122 | 28.9391 | 39.4509 | 42.9588 | 32.2936 | 24.8644 |
| | Accuracy | 71.1452 | 83.6298 | 87.3332 | 88.8488 | 91.7369 | 98.1041 | 98.6217 | 97.4221 | 89.5118 |
| Three 5.1581 | Fr Size | 10.8877 | 11.3894 | 11.7950 | 12.2746 | 12.8100 | 13.1560 | 15.3981 | 16.1796 | 16.7323 |
| | SNR | 11.6689 | 14.4418 | 18.0843 | 22.5843 | 28.9631 | 40.5176 | 42.4350 | 30.1566 | 22.5473 |
| | Accuracy | 81.6823 | 88.0950 | 88.6131 | 90.3372 | 91.8130 | 97.6215 | 98.6265 | 98.4385 | 81.1703 |
| Four 3.3196 | Fr Size | 8.6723 | 8.9856 | 9.3950 | 9.7304 | 12.3792 | 12.1521 | 12.8019 | 13.2287 | 14.1477 |
| | SNR | 10.5212 | 14.7670 | 19.1109 | 22.0829 | 29.9058 | 42.1800 | 51.0734 | 34.4307 | 24.6275 |
| | Accuracy | 73.6484 | 90.0787 | 93.6434 | 88.3316 | 94.8014 | 99.5448 | 99.0395 | 98.4060 | 88.6590 |
| Five 2.9423 | Fr Size | 11.1646 | 11.6298 | 12.0950 | 12.6381 | 13.0254 | 13.6579 | 15.9750 | 16.8354 | 17.4708 |
| | SNR | 10.0491 | 14.4815 | 18.1763 | 22.1886 | 28.4045 | 38.1060 | 46.0153 | 31.8556 | 23.7583 |
| | Accuracy | 70.3437 | 88.3372 | 89.0639 | 88.7544 | 90.0423 | 98.9302 | 98.4291 | 98.1957 | 85.5299 |
| Six 2.9731 | Fr Size | 7.1954 | 7.5433 | 7.8950 | 9.1852 | 9.7946 | 10.1444 | 10.4942 | 10.6056 | 11.1938 |
| | SNR | 10.0934 | 13.4992 | 17.5608 | 22.2304 | 29.4272 | 40.7459 | 46.3925 | 32.2664 | 23.7353 |
| | Accuracy | 70.6538 | 82.3451 | 86.0479 | 88.9216 | 93.2842 | 97.1603 | 98.1458 | 98.3459 | 85.4471 |
| Seven 3.963 | Fr Size | 14.0262 | 14.6346 | 15.2450 | 15.9092 | 16.4715 | 17.1713 | 17.7058 | 18.1469 | 18.9477 |
| | SNR | 10.4811 | 14.4564 | 18.0680 | 22.3551 | 29.4636 | 39.1440 | 42.7104 | 35.0012 | 26.0217 |
| | Accuracy | 73.3677 | 88.1840 | 88.5332 | 89.4204 | 93.3996 | 99.3798 | 98.1498 | 98.0034 | 93.6781 |
| Eight 7.5287 | Fr Size | 4.7031 | 5.7404 | 6.0950 | 6.4592 | 6.9946 | 5.8781 | 6.1673 | 12.2450 | 13.4092 |
| | SNR | 10.2369 | 15.4974 | 16.5973 | 21.9644 | 25.2904 | 26.2734 | 21.8446 | 30.0355 | 20.0605 |
| | Accuracy | 71.6583 | 94.5341 | 81.3268 | 87.8576 | 80.1706 | 96.0052 | 98.5047 | 98.0994 | 92.2178 |
| Nine 5.2752 | Fr Size | 12.9185 | 13.5529 | 14.0450 | 14.6371 | 15.1792 | 15.9165 | 16.5519 | 16.8354 | 19.6862 |
| | SNR | 11.5170 | 14.6621 | 19.1623 | 22.4067 | 29.4844 | 37.5927 | 38.1018 | 27.7750 | 22.9560 |
| | Accuracy | 80.6190 | 89.4388 | 93.8953 | 89.6268 | 93.4655 | 98.7188 | 97.3934 | 97.7700 | 82.6416 |

**TABLE 4:** SNR & Accuracy Test results for different frame size and frame overlap % and Window Size 260 for different samples at Real World Environment Noise

| Digit & SNR | Frame Overlap % | 20 | 25 | 30 | 35 | 40 | 45 | 50 | 55 | 60 |
|---|---|---|---|---|---|---|---|---|---|---|
| Zero 4.6704 | Fr Size | 7.7289 | 7.9583 | 8.3248 | 8.8450 | 9.0170 | 9.5270 | 9.8278 | 10.2128 | 10.4237 |
| | SNR | 10.5210 | 13.5585 | 17.4763 | 21.1063 | 26.8338 | 31.0465 | 30.3656 | 23.1879 | 22.1511 |
| | Accuracy | 84.1680 | 92.7069 | 85.6339 | 84.4252 | 85.0631 | 97.4070 | 96.1970 | 96.6073 | 79.7440 |
| One 2.532 | Fr Size | 10.3956 | 10.8519 | 11.3581 | 11.8200 | 12.1281 | 12.6687 | 13.1611 | 13.3702 | 14.3348 |
| | SNR | 10.2738 | 13.6145 | 16.8504 | 22.0041 | 28.3733 | 36.2857 | 40.0923 | 32.2709 | 24.3966 |
| | Accuracy | 82.1904 | 83.0485 | 92.5670 | 88.0164 | 89.9434 | 98.4571 | 97.4012 | 98.1314 | 87.8278 |
| Two 7.1607 | Fr Size | 11.1067 | 11.5463 | 12.0804 | 12.5200 | 12.9578 | 13.6354 | 3.9944 | 16.2119 | 7.1793 |
| | SNR | 10.8361 | 13.5814 | 18.6875 | 22.4416 | 29.9386 | 38.8200 | 41.9225 | 30.6552 | 24.8186 |
| | Accuracy | 86.6888 | 82.8465 | 91.5688 | 89.7664 | 94.9054 | 98.2860 | 99.3911 | 98.7690 | 89.3470 |
| Three 5.1581 | Fr Size | 10.4844 | 10.9676 | 11.3581 | 11.8200 | 12.3356 | 12.6687 | 4.8278 | 15.8961 | 6.1126 |
| | SNR | 10.0660 | 13.4298 | 17.6323 | 22.2393 | 29.4800 | 40.5937 | 25.0868 | 28.2966 | 22.0612 |
| | Accuracy | 80.5280 | 81.9218 | 86.3983 | 88.9572 | 93.4516 | 93.3655 | 98.6892 | 97.4008 | 79.4203 |
| Four 3.3196 | Fr Size | 8.4400 | 8.7685 | 9.1915 | 9.5450 | 11.2985 | 11.7020 | 2.3278 | 12.7387 | 3.6237 |
| | SNR | 10.5104 | 14.4805 | 18.0018 | 23.3613 | 29.5838 | 1.8883 | 9.7839 | 34.0240 | 44.8241 |
| | Accuracy | 84.0832 | 88.3311 | 88.2088 | 93.4452 | 93.7806 | 94.3431 | 97.3289 | 96.8648 | 96.3668 |
| Five 2.9423 | Fr Size | 10.8400 | 11.3148 | 11.6470 | 12.3450 | 12.7504 | 13.1520 | 15.3833 | 16.2119 | 16.8237 |
| | SNR | 10.4099 | 14.6564 | 19.1060 | 23.1816 | 28.4106 | 38.4336 | 44.6465 | 34.5082 | 26.6326 |
| | Accuracy | 83.2792 | 89.4040 | 93.6194 | 92.7264 | 90.0616 | 98.3973 | 97.3294 | 96.1721 | 95.8774 |
| Six 2.9423 | Fr Size | 10.8400 | 11.3148 | 11.6470 | 12.3450 | 12.7504 | 13.1520 | 15.3833 | 16.2119 | 16.8237 |
| | SNR | 10.4099 | 14.6564 | 19.1060 | 23.1816 | 28.4106 | 38.4336 | 44.6465 | 34.5082 | 26.6326 |
| | Accuracy | 83.2792 | 89.4040 | 93.6194 | 92.7264 | 90.0616 | 98.3973 | 97.3294 | 97.1721 | 95.8774 |
| Seven 3.963 | Fr Size | 13.6844 | 14.2083 | 14.8248 | 15.3200 | 16.0689 | 16.5354 | 17.3278 | 17.7906 | 18.6015 |
| | SNR | 11.4981 | 15.9106 | 17.2896 | 22.6451 | 28.7847 | 38.0097 | 41.0257 | 33.8041 | 23.9987 |
| | Accuracy | 91.9848 | 97.0547 | 84.7190 | 90.5804 | 91.2475 | 98.4223 | 99.4360 | 98.2711 | 86.3953 |
| Eight 7.5287 | Fr Size | 4.6178 | 5.5278 | 5.8693 | 6.3950 | 6.7356 | 5.6604 | 5.9389 | 11.7915 | 12.9126 |
| | SNR | 12.1827 | 16.1518 | 16.4825 | 20.9298 | 24.2040 | 26.6296 | 22.2510 | 26.2754 | 21.5904 |
| | Accuracy | 97.4616 | 98.5260 | 80.7643 | 83.7192 | 86.7267 | 96.2481 | 98.5072 | 97.9436 | 87.7254 |
| Nine 5.2752 | Fr Size | 12.6178 | 13.0509 | 13.6693 | 14.0950 | 14.8244 | 15.3270 | 15.9389 | 16.5276 | 19.3126 |
| | SNR | 9.6448 | 12.9770 | 17.0795 | 22.1698 | 29.4271 | 37.7130 | 38.8767 | 29.3890 | 20.1720 |
| | Accuracy | 87.1584 | 89.1597 | 83.6896 | 88.6792 | 93.2839 | 96.7399 | 98.7512 | 97.3503 | 82.6192 |

**TABLE 5:** SNR & Accuracy Test results for different frame size and frame overlap % and Window Size 270 for different samples at Real World Environment Noise